# Performance of a large language model-Artificial Intelligence-based chatbot for counseling patients with sexually transmitted infections and genital diseases

Nikhil Mehta[1], MD; Sithira Ambepitiya[2], MBBS; Thanveer Ahamad[2], MSc; Dinuka Wijesundara[2], MSc; Yudara Kularathne[2], MD

[1]Department of Dermatology and Venereology, University College of Medical Sciences and Guru Teg Bahadur Hospital, New Delhi, India

[2] HeHealth Inc., USA

**Corresponding Author:**

Nikhil Mehta, Department of Dermatology and Venereology, University College of Medical Sciences and Guru Teg Bahadur Hospital, New Delhi, India

E-mail id: nikhilmehtadermatology@gmail.com

**Abstract**

**Introduction:** The global burden of sexually transmitted infections (STIs) is rising, but there is a shortage of adequately trained counselors to address it. Current chatbots like ChatGPT are not tailored for handling STI-related concerns out of the box. We developed Otiz, an Artificial Intelligence-based (AI-based) chatbot platform designed specifically for STI detection and counseling, and assessed its performance.

**Methods:** Otiz employs a multi-agent system architecture based on GPT4-0613 (Generative Pre-trained Transformer model 4-0613), leveraging large language model (LLM) and Deterministic Finite Automaton (DFA) principles to provide contextually relevant, medically accurate, and empathetic responses. Its components include general STI information module, emotional recognition module, Acute Stress Disorder detection module, and psychotherapy module. Another question suggestion agent operates in parallel. Six common genital conditions- 4 STIs (anogenital warts, herpes, syphilis, urethritis/cervicitis) and 2 non-STIs (candidiasis, penile cancer) were evaluated using prompts mimicking patient language. Each prompt was independently graded by two venereologists conversing with Otiz as patient actors on six criteria: diagnostic accuracy, overall accuracy, relevance, correctness of information, comprehensibility, and empathy, using Numerical Rating Scale (NRS) ranging from 0 (poor performance) to 5 (excellent performance), and qualitative feedback.

**Results:** Twenty-three venereologists performed 60 evaluations of 30 prompts. Across STIs, Otiz scored highly on diagnostic accuracy (4.1-4.7), overall accuracy (4.3-4.6), correctness of information (5.0), comprehensibility (4.2-4.4), and empathy (4.5-4.8). However, relevance scores were lower (2.9-3.6), suggesting some redundancy. Diagnostic scores for non-STIs were lower (p=0.038). Inter-observer agreement was strong, with differences greater than 1 point on the NRS occurring in only 12.7% of paired evaluations (19 out of 150 pairs).

**Conclusions:** AI conversational agents like Otiz can provide accurate, correct, discrete, non-judgmental, readily accessible and easily understandable STI-related information in an empathetic manner. Further improvements are needed to enhance response relevance and reduce redundancy. For non-STIs, they could be useful for follow-up. With additional refinements, such chatbots can be integrated with sexual health services for a streamlined user experience, to alleviate the burden on healthcare systems.

**Manuscript**

**Introduction**

Sexually transmitted infections (STIs) have been on the rise globally in recent years.[1,2] According to the World Health Organization, there are more than one million new STI cases every day worldwide.[3,4] Despite this growing burden, healthcare systems often lack the infrastructure and workforce to provide adequate STI counseling and diagnosis, particularly in resource-limited settings.[5–7] Primary healthcare providers, already overstretched, often lack the time and specialized training needed to effectively counsel and diagnose patients with STI-related concerns.[7] The growing disease burden is not matched by an increase in medical infrastructure and workforce, posing a significant challenge. Addressing these systemic issues requires long-term reforms at multiple levels, but more immediate solutions are needed to bridge the gap in access to reliable STI information and support.[8]

STIs and genital diseases carry significant social stigma, leading to hesitancy among patients in discussing these sensitive issues with healthcare providers.[9] This reluctance poses a major barrier to timely diagnosis and treatment. Fear of stigma and judgment has been shown to be a significant reason for patients delaying or avoiding seeking STI-related care.[10] While there is a wealth of STI-related information available online, much of it is unreliable or presented in a manner that is not easily comprehensible to the general public.[11] Even reputable sources like government health websites often use medical jargon that can be difficult for patients to understand.

Artificial intelligence (AI) chatbots present a promising solution to these challenges. AI research in STIs has mostly focused on image diagnosis through computer vision algorithms, but conversational agents also have a significant use case.[12] By providing a confidential, accessible, and user-friendly platform for STI-related queries, chatbots can help patients obtain accurate information and guidance without the fear of stigma or judgment.[13] However, current large language models (LLMs) are not specifically designed to handle the nuances of sexual health concerns, and might censor some words or not accept them as inputs.

To address this unmet need, we developed Otiz, an AI-based chatbot platform tailored specifically for STI counseling and diagnosis. Otiz leverages LLM and Deterministic Finite Automaton (DFA) principles to provide contextually relevant, medically accurate, and empathetic responses to user queries.[14,15]

**Aims and Objectives**

The aims of this study were to develop an interactive, empathetic, and accurate AI-based chatbot, and to evaluate its performance in providing STI counseling and diagnosis.

**Methods**

I. System Architecture

Otiz is based on the GPT-4 language model (version GPT4-0613). It employs a multi-agent system architecture. Four modules, comprising numerous text prompt statements, are overlaid on GPT-4.

1. General STI information module- replicates the diagnostic process of an expert venereologist through systematic analysis of primary complaints, generation of differential diagnoses, sequential formulation of follow-up questions, provision of diagnoses or differential diagnoses, and detailed information on diseases, investigations, and treatments. It employs advanced algorithms for symptom-based detection of STIs.[13,16] The module also offers guidance on seeking medical attention and provides resources for further information.

2. Emotional recognition module- activated post-diagnosis, it assesses emotional states through text analysis, identifies emotions ranging from anxiety to relief, and adapts subsequent responses for empathetic care. It utilizes sentiment analysis techniques to identify the user's mood and adapt the chatbot's responses accordingly.[17]

3. Acute Stress Disorder (ASD) detection module- employs a series of carefully crafted questions and prompts to assess the user's psychological state and determine the likelihood of ASD. If ASD is detected, the module provides targeted support and resources to help the user cope with their diagnosis.

4. Psychotherapy module- offers basic psychotherapeutic support, such as guided breathing exercises and relaxation techniques, progressive muscle relaxation techniques, and cognitive restructuring strategies, to help users manage stress and anxiety related to their sexual health concerns. This module aims to bridge the gap in patient-physician communication regarding mental health and provide immediate support to users in need.

Deterministic Finite Automaton (DFA) principle and conversational flow- DFA is a computational model that defines systems through a limited set of states with predefined transitions, allowing for precise control over user interactions and chatbot responses, ensuring that the conversation progresses logically and coherently.[14] This design is critical for ensuring both the precision of medical diagnostics and empathetic patient interactions.[15] It allows for precise control over user interactions and chatbot responses, and enables the implementation of context-specific prompts and questions, ensuring that the chatbot's responses are always relevant to the user's current situation. This approach helps maintain user engagement and promotes a more natural, human-like conversation. The seamless transitions between modules enable Otiz to adapt to the user's needs in real-time, mirroring the dynamic nature of human healthcare provider interactions while maintaining a structured and comprehensive approach to care.

At any point during the interaction with the Emotional Recognition, Psychotherapy, or ASD Detection Modules, the system can transition back to the STI Information Module if the user expresses a need for more medical information or has additional health-related questions. Throughout these transitions, the Question Suggestion Agent operates in parallel, providing context-aware prompts to guide the conversation and ensure comprehensive coverage of relevant topics. The users can either select suggested questions or type independently.

II.  <u>Prompt engineering</u>

The system utilizes a sophisticated, multi-layered prompting technique that imbues the LLM with the persona of an expert venereologist while simultaneously providing a framework for medical reasoning and decision-making. The system prompt begins by defining Otiz as an 'Expert venereologist physician.' This persona establishment is crucial as it sets the tone for all subsequent interactions and decision-making processes. The prompt further refines this persona by specifying character traits such as being "a consummate professional," "witty," "warm and kind," which guide the LLM's communication style. The system prompt incorporates metacognitive frameworks that instruct the LLM to engage in step-by-step analysis of symptoms, consideration of multiple diagnoses, and systematic questioning. The system begins by assessing the user's communication style and emotional state, leveraging the adaptive communication guidance embedded in the prompt. This initial assessment informs the tone and complexity of subsequent interactions. The system asks follow-up questions sequentially, as instructed by the prompt. This approach mimics the logical progression of a medical consultation, allowing for a thorough exploration of the user's condition while maintaining a natural conversational flow. Findings are presented in a manner tailored to the user's comprehension level, as determined by the initial communication style assessment. The prompt ensures that Otiz always provides a diagnosis or differential diagnosis, avoiding vague or generalized statements.

A unique aspect of Otiz's prompt engineering is the inclusion of adaptive communication strategies. The prompt instructs the LLM to assess the user's tone and language on a scale of 1 to 10 and modify its responses accordingly, while never falling below a professionalism level of 4. Other prompts embed ethical considerations directly into the agent's decision-making process. For example, it explicitly instructs Otiz to always recommend professional medical evaluation. Rather than relying on pre-programmed responses, the prompt guides the LLM to apply its vast knowledge base in a structured, clinically relevant manner. It provides specific instructions for addressing various aspects of sexual health, from low sexual desire to partner concerns, ensuring comprehensive coverage of relevant topics.

By integrating multiple functionalities within a single agent through careful prompt design, Otiz is designed to offer a natural, conversational experience that closely mimics interaction with a human healthcare provider, while maintaining consistent professionalism and ethical standards.

### III. Evaluation of utility of Otiz

To evaluate Otiz's performance, six common genital conditions- 4 common STIs (anogenital warts, herpes, primary chancre of syphilis, urethritis/cervicitis) and 2 non-STIs (penile candidiasis, penile cancer) were evaluated. The 2 non-STI conditions were included to see the additional impact of STI information module, compared to other genital conditions. For each condition, five unique initiating text prompts, of one or two lines each were created, mimicking patient language, descriptions, and phrasing. The prompts were designed to represent a diverse range of manifestations and presentations and included both straightforward and more complex cases (**Supplement 1**) by 2 expert venereologists with more than 2 years of experience of store-and-forward telemedicine services.

A different set of 23 venereologists were given various initiating prompts, and were asked to interact with Otiz as patient actors. They were instructed to converse with the chatbot as patients converse with them, and to ask questions which the patients commonly ask them, in their language. Each venereologist was assigned a maximum of three prompts to evaluate, to avoid bias. It was also ensured that each prompt was independently assessed by two different venereologists, to enable comparison. This amounted to 60 evaluations of 30 unique prompts.

After completing the interaction, the venereologists graded the chatbot's responses on six criteria using a 6-point Numerical Rating Scale ranging from 0 (indicating poor performance), to 5 (indicating excellent performance):

1. Diagnostic accuracy: The ability of the chatbot to suggest the correct diagnosis or differentials based on the provided symptoms.

2. Overall accuracy: The general accuracy of the chatbot's responses in relation to the user's queries.

3. Relevance: The relevance of the chatbot's responses to the user's specific concerns and the absence of unnecessary or redundant information.

4. Correctness of information: The accuracy and reliability of the medical information provided by the chatbot.

5. Comprehensibility: The clarity and understandability of the chatbot's responses for a lay audience.

6. Empathy: The chatbot's ability to demonstrate empathy and provide emotionally supportive responses.

The venereologists were also asked to provide qualitative feedback on the chatbot's strengths and weaknesses, as well as suggestions for improvement.

IV. Data Analysis

The data collected from the venereologists' evaluations were analyzed using descriptive statistics. For each criterion, the mean and median scores were calculated across all prompts within each disease class to compare between different STI-related conditions, and with non-STI conditions. Mean NRS values between STIs and non-STIs were compared using Wilcoxon signed-rank test for paired non-parametric data (p<0.05 considered significant). Inter-observer agreement between STI specialists for the same prompt was calculated by pairs with more than 1 point difference in NRS.

The qualitative feedback provided by the venereologists was analyzed using thematic analysis. Common themes and patterns were identified in the feedback, providing insights into the chatbot's strengths, weaknesses, and areas for improvement.

**Results**

Quantitative analysis

The mean and median NRS scores for each evaluation criterion across the six disease classes are presented in **Table 1**. Except for relevance and diagnostic accuracy, all rest 4 parameters had mean and median scores above 4. Diagnostic scores were also above 4 for STIs but not for non-STIs. There was significant difference between diagnostic accuracy of STI vs non-STIs (p=0.038). The proportion of responses varying by more than 1 point in NRS by 2 STI specialists was low, 19/150 pairs (12.7%), max for warts (7) and discharge/proctitis (6), demonstrating good inter-observer agreement.

Qualitative Analysis

The qualitative feedback provided by the venereologists revealed several strengths and weaknesses of the Otiz chatbot. The most frequently mentioned strengths included:

- Reliable medical information without any misinformation (n=13, 56.52%)
- Empathetic and supportive tone (n=7, 30.43%)
- Clear and easily understandable language (n=5, 21.74%)
- Ability to provide relevant resources and recommendations (n=3, 13.04%)

The main weaknesses identified by the venereologists were:

- Redundancy in some responses, with repetition of information (n=13, 56.52%)
- Occasional inclusion of irrelevant or unnecessary details (n=10, 43.48%)
- Slow response rate (n=8, 34.78%)
- Excessive focus on mental health, with difficulty in going back on focusing on the case (n=7, 30.43%)
- Limited ability to handle complex or atypical cases (n=2, 8.7%)

The venereologists also provided suggestions for improving the chatbot, such as:

1. Refining the algorithms to reduce redundancy and improve the relevance of responses
2. Expanding the knowledge base to cover a wider range of presentations and edge cases
3. Incorporating more visuals and multimedia content to enhance user engagement and understanding
4. Enabling easy and rapid switching to topics/areas of patients' interest, reducing the focus on mental health

**Discussion**

This is the first proof-of-concept of an STI conversational agent in the world, tested by a decent number of doctors performing as patient actors. People want discrete, non-judgmental, empathetic provider instantly accessible anytime of the day, who can protect them from misinformation. All of this is essentially possible by a non-human provider. The evaluation of the Otiz chatbot by venereologists demonstrated its strong performance in providing accurate, comprehensible, and empathetic responses to users seeking information and support related to STIs. The chatbot's high scores on detection accuracy, overall accuracy, and correctness of information suggest that it can serve as a reliable source of medical information for users. These findings are in line with previous studies that have highlighted the potential of AI-powered chatbots in delivering accurate and reliable health information.[18,19] Similar platforms exist for mental health disorders, and are US FDA approved,[20,21] Organisation for the Review of Care and Health Apps (ORCHA) approved, recommended and deployed by UK's National Health Service (NHS).[22] They have good acceptance in the targeted users.[23] STI counselling is an equally good use case for a chatbot, as it too carries stigma and requires discretion, and can be similarly integrated to other services, at least for follow-up. However, Otiz is not a replacement for doctor or medical care. It needs to be further studied in real-life/practical settings as a supplemental tool, if it is able to save providers' time in resource-limited settings by assisting in patient counseling after the initial visit, enabling them to see more patients in the same time, and improve patient satisfaction.

The chatbot's excellent comprehensibility and empathy scores highlight its potential to bridge the gap in access to sexual health information and support, particularly for individuals who may be hesitant to discuss these sensitive topics with healthcare providers.[13] By providing clear, easily understandable information in a supportive and non-judgmental manner, Otiz can help reduce the barriers to seeking sexual health advice and encourage users to take proactive steps in managing their health, similar to other chatbots.[24,25]

The best strength of the chatbot in the study was a complete lack of any misinformation, in line with 'first do no harm' principles. Non-STIs had lower score for diagnosis obviously because the STI diagnostic module wasn't very helpful for them, they had more edge cases, and less uniform presentation.

The use of DFA principles and a modular architecture in Otiz's design is a novel approach that sets it apart from other chatbots in this domain. The DFA framework allows for more precise control over the conversational flow and enables the chatbot to provide more context-specific and relevant responses. This approach addresses some of the limitations of previous chatbots, such as lack of coherence and adaptability to user needs, which have been identified as important challenges in the development of health chatbots.[26,27]

However, the relatively lower scores on relevance indicate that there is room for improvement in the chatbot's algorithms to ensure that responses are more concise and targeted to the user's specific concerns. Reducing redundancy and minimizing the inclusion of irrelevant information can enhance the user experience and make the chatbot more engaging and efficient.[17,28]

The qualitative feedback provided by the venereologists offers valuable insights into the strengths and weaknesses of the chatbot, as well as potential areas for future development. Refining the algorithms to improve the relevance of responses, expanding the knowledge base to cover a wider range of cases, and incorporating more visuals and multimedia content are all promising avenues for enhancing the chatbot's functionality and user experience.[29,30]

**Limitations and Future Work**

One limitation of this study is that the evaluation was conducted by venereologists mimicking patient interactions, rather than by actual patients. While this approach allowed for a more controlled and standardized evaluation, it may not fully capture the diversity of user needs and preferences in real-world settings. Future studies should involve actual patients to assess the chatbot's performance and user experience in more ecologically valid contexts.[28,31]

Another limitation is that the study focused on a specific set of STI-related conditions and may not represent the full range of sexual health concerns that users may have. The prompt designing was done by two experts based on their experience. Using proportional thematic analysis of actual presenting complaints of the teledermatology patients, and including more experts could have made it more representative. Incorporation of more edge cases and ambiguous queries simulating more real-life challenges will be a better assessment of robustness, and should be done while assessing future versions or other such chatbots. Expanding the chatbot's knowledge base to cover a broader spectrum of sexual health topics would enhance its utility and reach.[18,19]

Future work should also focus on integrating the chatbot with other sexual health services, such as testing and treatment facilities, to provide a more comprehensive and streamlined user experience.[32] Incorporating user feedback and machine learning techniques to continuously improve the chatbot's performance and adaptability to user needs is another important avenue for future development.[33] Blinded trials comparing chatbot interactions with those of human counselors, and non-specific chatbots, using randomized anonymized transcripts of patients' interactions, can be reviewed by blinded assessors with validated scores to better quantify the additional impact of an STI-specific chatbot.[34] Other non-STI genital diseases should also be covered by a separate module in a single chatbot, for an integrated genital disease counseling conversational agent.

**Conclusion**

Otiz serves as a reliable and accessible resource for individuals seeking STI-related information and support. The findings of this study could pave the way for the integration of AI chatbots into sexual health services, helping to bridge the gap in access to timely, accurate, and empathetic STI counseling and diagnosis. Successfully demonstrating the effectiveness of AI chatbots in this context could have significant implications for public health, particularly in resource-limited settings where access to specialized STI care is limited.

**Competing interests:** Otiz chatbot is developed by HeHealth Inc. SA, TA, DW and YK are employees at HeHealth Inc. NM does not have any relevant competing interests.

**Tables**

Table 1. Mean and median scores for evaluation criteria by disease class

| Disease Class | Diagnostic Accuracy | Overall Accuracy | Relevance | Correctness of Information | Comprehensibility | Empathy |
|---|---|---|---|---|---|---|
| Anogenital warts | 4.2 ± 0.92 (4) | 4.3 ± 0.67 (4) | 3.6 ± 0.97 (4) | 5.0 ± 0.0 (5) | 4.4 ± 0.84 (5) | 4.5 ± 0.53 (4.5) |
| Anogenital herpes | 4.7 ± 0.48 (5) | 4.5 ± 0.53 (4.5) | 3.5 ± 0.97 (3) | 5.0 ± 0.0 (5) | 4.4 ± 0.52 (4) | 4.7 ± 0.48 (5) |
| Gonorrhea/Chlamydia/UTI | 4.6 ± 0.70 (5) | 4.6 ± 0.70 (5) | 3 ± 1.05 (3) | 5.0 ± 0.0 (5) | 4.3 ± 1.06 (5) | 4.8 ± 0.42 (5) |
| Primary syphilis | 4.1 ± 0.74 (4) | 4.3 ± 1.06 (5) | 2.9 ± 0.74 (3) | 5.0 ± 0.0 (5) | 4.2 ± 0.92 (4.5) | 4.7 ± 0.48 (5) |
| Penile cancer | 2.9 ± 1.5 (3) | 3.6 ± 0.63 (4) | 3.3 ± 1.38 (3) | 5.0 ± 0.0 (5) | 4.4 ± 0.75 (5) | 4.8 ± 0.41 (5) |
| Penile candidiasis | 3.5 ± 1.72 (4) | 4 ± 0.94 (4) | 3.2 ± 0.79 (3) | 5.0 ± 0.0 (5) | 4.2 ± 1.03 (4.5) | 4.7 ± 0.48 (5) |